\documentclass{article}

\PassOptionsToPackage{numbers,compress}{natbib}
\usepackage[preprint]{neurips_2026}

\usepackage[utf8]{inputenc}
\usepackage[T1]{fontenc}

\usepackage{amsmath}
\usepackage{amssymb}
\usepackage{amsfonts}
\usepackage{nicefrac}

\usepackage{graphicx}
\usepackage{subcaption}
\usepackage{wrapfig}

\usepackage{booktabs}
\usepackage{makecell}
\usepackage{tabularx}

\usepackage{microtype}
\usepackage{xcolor}
\usepackage{pifont}
\usepackage{placeins}

\usepackage{url}
\usepackage[
    colorlinks=true,
    linkcolor=black,
    citecolor=black,
    urlcolor=blue
]{hyperref}

\usepackage[most]{tcolorbox}

\usepackage{fancyhdr}

\usepackage{colortbl}

\usepackage{fontawesome5}

\definecolor{AtomWorldRow}{HTML}{EAF2F8}
\definecolor{TableHeader}{HTML}{F4F6F8}
\definecolor{GroupHeader}{HTML}{EEF3F7}

\newcommand{\cmark}{\ding{51}}
\newcommand{\xmark}{\ding{55}}

\definecolor{AtomPurple}{HTML}{7B35D5}
\definecolor{AtomBlue}{HTML}{4E79C6}
\definecolor{AtomCyan}{HTML}{31BFA9}

\definecolor{FrontBoxBlue}{HTML}{F0F4F8}

\makeatletter

\renewcommand{\@toptitlebar}{}
\renewcommand{\@bottomtitlebar}{}

\renewcommand{\@notice}{}

\makeatother

\AtBeginDocument{
  \newgeometry{
    letterpaper,
    left=0.78in,
    right=0.78in,
    top=1.05in,
    bottom=0.72in,
    headheight=0.58in,
    headsep=0.12in,
    footskip=24pt
  }
}

\newtcolorbox{frontmatterbox}{
    enhanced,
    width=\textwidth,
    colback=FrontBoxBlue,
    colframe=FrontBoxBlue,
    boxrule=0pt,
    arc=12pt,
    outer arc=12pt,
    left=8mm,
    right=8mm,
    top=7mm,
    bottom=7mm,
    before skip=0pt,
    after skip=8mm
}

\fancypagestyle{airfirstpage}{
    \fancyhf{}

    \fancyhead[L]{%
        \raisebox{-0.05in}{%
            \includegraphics[
                height=0.48in,
                keepaspectratio
            ]{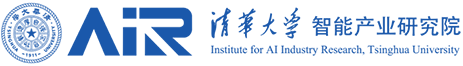}%
        }%
    }

    \fancyfoot{}

}

\begin{document}

\thispagestyle{airfirstpage}


\begin{frontmatterbox}


{\raggedright
\fontsize{22.5}{25.5}\selectfont
\bfseries

{\color{AtomPurple}Atom}%
{\color{AtomBlue}World}%
{\color{AtomCyan}-Mem}:
Memory-Restored World States for Long-Horizon Atomistic Evolution

\par
}

\vspace{0.38cm}


{\raggedright
\fontsize{10.8}{13.2}\selectfont
\bfseries

Tian Luo$^{1,5\dagger}$,
Ruge Zhang$^{2,3,5,\dagger}$,
Haozhi Han$^{4,5}$,
Yifeng Chen$^{4}$,
Yunquan Zhang$^{2,3}$,
Ting Cao$^{5}$,
Yunxin Liu$^{5}$,
Kun Li$^{5,\ddagger}$

\par
}

\vspace{0.22cm}


{\raggedright
\fontsize{9.7}{11.9}\selectfont
\normalfont

$^{1}$Sichuan University, Chengdu, China\\
$^{2}$Institute of Computing Technology, Chinese Academy of Sciences, Beijing, China\\
$^{3}$University of Chinese Academy of Sciences, Beijing, China\\
$^{4}$School of Computer Science, Peking University, Beijing, China\\
$^{5}$Institute for AI Industry Research (AIR), Tsinghua University, Beijing, China\\[2pt]

$^{\dagger}$Equal contribution.
\qquad
$^{\ddagger}$Corresponding author.\\[2pt]
\textit{This work was supported by Tecorigin.}

\par
}

\vspace{0.44cm}


{\fontsize{9.8}{12.1}\selectfont
\normalfont

High-fidelity atomistic evolution over long timescales requires more than observing the current crystal configuration. Instantaneous atomistic snapshots are often incomplete: locally similar configurations can correspond to different hidden dynamical contexts, future event preferences, and waiting-time scales. We argue that this \emph{snapshot ambiguity} makes long-horizon atomistic evolution fundamentally a memory-based world-state restoration problem. To address this, we introduce \textbf{AtomWorld-Mem}, a memory-restored atomistic world model that recovers the latent world state missing from instantaneous crystal snapshots. AtomWorld-Mem treats the evolving alloy as an \emph{AtomWorld}: spatial encoders write multi-scale atomistic keyframes from dense local topology and sparse long-range defect context, while short-term event memory and long-term structural memory integrate these keyframes across time to restore a future-predictive evolutionary state. The restored state is used to prioritize legal vacancy-mediated events under single-event Kinetic Monte Carlo (KMC) constraints, while event legality, physical execution, and residence-time updates remain governed by the underlying simulator. Empirically, AtomWorld-Mem improves long-horizon atomistic progress under fixed microscopic event budgets while maintaining high-fidelity evolution across energetic, structural, and vacancy-transport observables. It further transfers zero-shot across diverse unseen alloy--temperature AtomWorlds, suggesting that the learned memory-restoration mechanism captures reusable principles of hidden-state inference rather than a system-specific local energy heuristic. These results position memory-restored world-state modeling as a promising route toward efficient, physically grounded, and transferable atomistic evolution.

\par
}

\vspace{0.34cm}


{\raggedright
\fontsize{9.2}{11.2}\selectfont

\textbf{Email:}
\href{mailto:likun@air.tsinghua.edu.cn}{likun@air.tsinghua.edu.cn}%
\hspace{8em}%
\textbf{Project Page:}
\href{https://atomworld-mem.github.io/}{https://atomworld-mem.github.io/}

\par
}

\end{frontmatterbox}


\section{Introduction}
Simulating long-horizon atomistic evolution at microscopic fidelity is crucial across scientific domains, from radiation damage to catalysis and phase transformations~\cite{english2012radiation, pineda2022kinetic, martin2005kinetic}. Existing long-timescale atomistic methods advance a system through admissible local transitions and update physical time according to transition rates~\cite{uberuaga2020computational, voter2007introduction}. This rate-driven principle preserves physical consistency, but also exposes a fundamental blind spot: instantaneous rates do not reveal whether an event is a reversible local fluctuation or a transition that drives structural evolution~\cite{kratzer2009monte, stamatakis2012unraveling, reuter2012first}. Consequently, simulation effort can be repeatedly consumed within locally metastable regions, delaying access to rare but consequential transitions that shape macroscopic structure under practical budgets~\cite{stamatakis2012unraveling, reuter2012first}.

\begin{figure}[htbp]
    \centering
    \includegraphics[width=\linewidth]{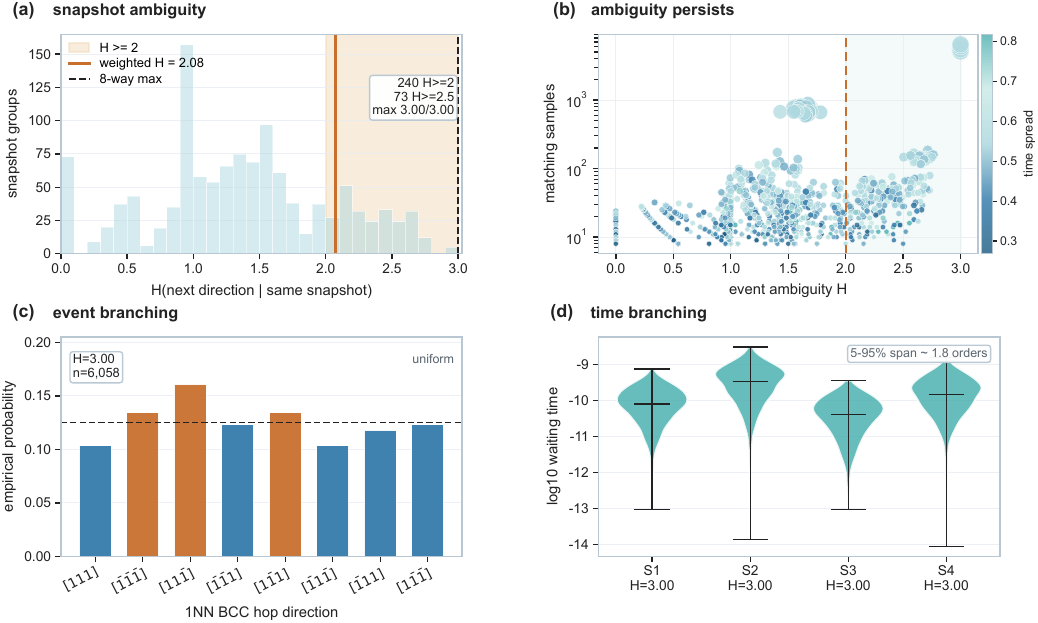}
    \caption{\textbf{Snapshot ambiguity in atomistic evolution.} Empirical analysis shows that identical local spatial observations can branch into different future trajectories and exhibit drastically different waiting times. This indicates that a single static snapshot is not a sufficient predictive state for long-horizon atomistic evolution.}
    \label{fig:snapshot_ambiguity}
    \vspace{-1.2em}
\end{figure}

A tempting direction is to make event selection aware of long-horizon consequences while remaining within the admissible event space. Yet this is ill-posed from the current spatial snapshot alone due to \emph{snapshot ambiguity}: similar local configurations can branch into distinct future event distributions and markedly different waiting-time scales, as shown in Figure~\ref{fig:snapshot_ambiguity}. In learning terms, an instantaneous atomic snapshot is not a sufficient statistic for the future~\cite{kaelbling1998planning, eberhard2025partially}; it aliases multiple hidden dynamical contexts into the same observation.
What is missing is not another local descriptor, but the world state of the evolving atomic system: a latent dynamical representation that summarizes the hidden context needed for future event reasoning~\cite{wang2018deepmd, wang2024latent, nasim2024dynamically}. The central challenge is therefore to restore this world state from partial observations and temporal memory, enabling long-horizon event reasoning beyond instantaneous local rates.

To address this challenge, we model atomistic evolution as an evolving atomic world, where memory bridges instantaneous snapshots and long-horizon dynamics to recover the latent world state that snapshots alone cannot reveal. We introduce \textbf{AtomWorld-Mem}, a memory-restored atomistic world model that reconstructs this state across time, transforming snapshot-limited transition scoring into memory-grounded reasoning over atomistic evolution.

Concretely, AtomWorld-Mem implements this memory-restored world-state formulation through four coupled components: a \textit{Keyframe Writer}, a \textit{Latent-State Memory}, a\textit{ Legal Event Scorer}, and a \textit{Physics-Guided Policy Learner}. The Keyframe Writer records compact, high-information keyframes from multi-scale atomistic observations, combining dense short-range local topology for immediate admissible transitions with sparse long-range defect context for slower structural evolution. The Latent-State Memory performs dual-scale temporal reasoning:\textbf{ short-term memory} captures recent action history and local transition competition, while\textbf{ long-term memory} accumulates structural biases and hidden defect-evolution trends across the trajectory. Together, these memories restore a latent world state from partial observations, forming a dynamically updated representation of the current crystal evolutionary context. The Legal Event Scorer evaluates admissible transitions under the restored state, while the Physics-Guided Policy Learner optimizes their prioritization under single-event KMC execution. By operating within the admissible event space and preserving physical time semantics, AtomWorld-Mem enables efficient, time-consistent evolution without sacrificing physical fidelity.

Empirically, the results support this state-restoration view. AtomWorld-Mem reaches up
to 420$\times$ fixed-budget acceleration under matched legal-event budgets, showing
that the restored world state enables long-horizon event reasoning beyond instantaneous
local rates. The accelerated rollouts remain aligned with reference energetic,
structural, and vacancy-transport trends in physical-time space. Memory ablations reveal
complementary roles: short-term memory supports recent-event reasoning and vacancy
arrival, whereas long-term memory supports structural-context accumulation and rollout
fidelity. Zero-shot transfer across unseen alloy--temperature AtomWorlds further suggests
that AtomWorld-Mem learns reusable hidden-state inference rather than a system-specific
local energy heuristic.

\section{Problem Formulation}

We model long-horizon atomistic evolution as an \emph{AtomWorld}: a partially observed
atomistic world evolving under single-event KMC constraints. At step $t$, the complete
microscopic configuration $x_t\in\mathcal{X}$ determines the legal event set, physical
transition rates, and subsequent atomistic evolution. The learned model does not observe
this complete state; it receives only structured partial observations and must infer the
hidden evolutionary context needed for long-horizon decision making.

Under the single-event KMC constraint, each decision executes exactly one physically
admissible vacancy-mediated exchange. For compactness, we write the legal event set as
\begin{equation}
\mathcal{A}_t=\mathcal{A}^{\mathrm{legal}}(x_t).
\label{eq:legal_event_set}
\end{equation}
A legal event $a_t\in\mathcal{A}_t$ updates the AtomWorld by
\begin{equation}
x_{t+1}=f(x_t,a_t).
\label{eq:kmc_transition}
\end{equation}
Each legal event has a physically defined base rate $r_t^{\mathrm{base}}(a)$. Conventional
rejection-free KMC samples events proportionally to these rates,
\begin{equation}
p_{\mathrm{KMC}}(a\mid x_t)
=
\frac{r_t^{\mathrm{base}}(a)}
{\sum_{a'\in\mathcal{A}_t} r_t^{\mathrm{base}}(a')}.
\label{eq:kmc_policy}
\end{equation}
Eq.~\eqref{eq:kmc_transition} defines the physically admissible state update, while
Eq.~\eqref{eq:kmc_policy} defines the passive rate-proportional event-selection rule.
This preserves event legality and micro-kinetic consistency, but does not explicitly
infer which admissible event is most consequential for long-horizon structural evolution.

\subsection{Partial Observation and Legal Action Space}

The complete configuration $x_t$ determines legal events, base rates, and physical
transitions, but the learned model only receives
\begin{equation}
o_t=\Omega(x_t)=
\bigl(G_t^{\mathrm{local}},\,S_t^{\mathrm{def}}\bigr),
\label{eq:partial_obs}
\end{equation}
where $G_t^{\mathrm{local}}$ is a vacancy-centered local graph over structured
1NN--4NN shells, and $S_t^{\mathrm{def}}$ is a set of sparse defect tokens encoding
longer-range defect context. AtomWorld-Mem is vacancy-centered in representation, but
the control problem remains event-level: the policy does not generate unconstrained
atomic moves, and all decisions are restricted to the legal set $\mathcal{A}_t$ defined
in Eq.~\eqref{eq:legal_event_set}. Together, Eqs.~\eqref{eq:legal_event_set} and
\eqref{eq:partial_obs} define the key partial-observation setting: the model acts only
through legal KMC events, but must do so from incomplete observations.

\subsection{Memory-Based State Restoration}

A single local snapshot is generally not a sufficient predictive state for long-horizon
atomistic evolution. Similar local observations may correspond to different hidden
dynamical contexts, leading to different future event preferences, waiting-time scales,
and structural trajectories. We therefore treat atomistic decision making as latent-state
restoration under partial observability.

AtomWorld-Mem writes a compact spatial keyframe
\begin{equation}
k_t=\psi(o_t),
\label{eq:keyframe_write}
\end{equation}
and infers a memory-restored latent state from keyframes and past actions,
\begin{equation}
z_t=\Phi(k_{\leq t},a_{<t}).
\label{eq:state_restoration}
\end{equation}
Here, $z_t$ represents hidden dynamical context that is missing from the current snapshot
but useful for future atomistic evolution. It does not denote an unconstrained generative
simulator. The policy acts on this restored state while remaining constrained to the legal
KMC event set:
\begin{equation}
a_t\sim
\pi_\theta(\cdot\mid z_t,\mathcal{A}_t),
\qquad
\operatorname{supp}\pi_\theta(\cdot\mid z_t,\mathcal{A}_t)
\subseteq
\mathcal{A}_t .
\label{eq:restored_policy}
\end{equation}
Thus, Eqs.~\eqref{eq:keyframe_write}--\eqref{eq:restored_policy} formalize the central
state-restoration view of AtomWorld-Mem: spatial observations are written into keyframes,
memory restores the latent evolutionary state, and the policy prioritizes only legal events.

\subsection{Training Objective under KMC Constraints}

The goal is to improve long-horizon atomistic progress under a fixed legal-event budget $T$.
We define the immediate energy drop and shaped training reward as
\begin{equation}
\Delta E_t = E(x_t)-E(x_{t+1}), 
\qquad 
r_t^{\mathrm{train}} = \lambda_E \Delta E_t + r_t^{\mathrm{shape}},
\label{eq:training_reward}
\end{equation}
and optimize
\begin{equation}
J(\pi_\theta)=
\mathbb{E}_{\pi_\theta}\!\left[
\sum_{t=0}^{T-1}\gamma^t r_t^{\mathrm{train}}
\right].
\label{eq:return_objective}
\end{equation}
Learning changes the decision rule for prioritizing legal events under partial observability;
event legality, physical execution, and time advancement remain governed by the underlying KMC process.



\section{AtomWorld-Mem}


AtomWorld-Mem implements the formulation in Section~2 with four modules:
a Keyframe Writer, Latent-State Memory, Legal Event Scorer, and Physics-Guided
Policy Learner.

\subsection{Architecture and Formulation}

AtomWorld-Mem follows a memory-centered principle: spatial encoders do not directly
define the predictive state. Instead, they write high-information atomistic keyframes,
and memory restores the missing evolutionary state by integrating these keyframes with
past legal events. Space provides the instantaneous atomistic context, while memory
converts partial observations into a future-predictive latent world state.

\paragraph{Keyframe Writer.}
The Keyframe Writer converts the partial observation $o_t$ in
Eq.~\eqref{eq:partial_obs} into a compact spatial keyframe. It uses two complementary
spatial pathways:
\begin{equation}
h_t^{\mathrm{dense}}=f_{\mathrm{gnn}}(G_t^{\mathrm{local}}),
\qquad
h_t^{\mathrm{sparse}}=f_{\mathrm{attn}}(S_t^{\mathrm{def}}),
\qquad
k_t=\phi_{\mathrm{fuse}}([h_t^{\mathrm{dense}}\|h_t^{\mathrm{sparse}}]).
\label{eq:keyframe_writer}
\end{equation}
The dense pathway captures short-range topology that determines immediate legal
transitions, while the sparse pathway captures long-range defect context that shapes
slower structural evolution. The keyframe $k_t$ is not treated as a complete state; it
is a spatial write into memory that must be interpreted together with past evolution.

\paragraph{Latent-State Memory.}
The Latent-State Memory restores hidden evolutionary context by integrating spatial
keyframes with previous legal events, following the state-restoration view in
Eq.~\eqref{eq:state_restoration}. We decompose this restoration into two temporal
scales. The short-term event memory summarizes recent action history and local
transition competition:
\begin{equation}
c_t^{\mathrm{short}}
=
\phi_{\mathrm{hist}}\!\left(\phi_{\mathrm{act}}(a_{t-H:t-1})\right).
\label{eq:short_memory}
\end{equation}
It captures recent vacancy motion, local reversibility, repeated back-and-forth
exchanges, and short-horizon competition among nearby legal events. The long-term
structural memory accumulates slower evolutionary biases:
\begin{equation}
m_t^{\mathrm{long}}
=
f_{\mathrm{rnn}}\!\left(m_{t-1}^{\mathrm{long}},[k_t\|c_t^{\mathrm{short}}]\right).
\label{eq:long_memory}
\end{equation}
This recurrent state retains hidden defect-evolution trends, aggregation tendency,
vacancy--solute interaction patterns, and structural biases over many KMC steps. The
restored world state is obtained by combining instantaneous space, short-term event
context, and long-term structural memory:
\begin{equation}
z_t
=
\phi_{\mathrm{mem}}\!\left([k_t\|c_t^{\mathrm{short}}\|m_t^{\mathrm{long}}]\right).
\label{eq:restored_state}
\end{equation}
Thus, $z_t$ is not merely an encoding of the current observation; it is a
memory-restored latent state used for both future event preference and time-scale
reasoning.

\paragraph{Legal Event Scorer.}
Given the restored state $z_t$, AtomWorld-Mem scores each legal event
$a\in\mathcal{A}^{\mathrm{legal}}(x_t)$. For each event, we construct
\begin{equation}
\xi_t(a)=\bigl(\tilde{G}_t(a),d(a),\mu_t(a)\bigr),
\label{eq:event_descriptor}
\end{equation}
where $\tilde{G}_t(a)$ is event-local structure, $d(a)$ is the event direction, and
$\mu_t(a)$ denotes auxiliary energetic or kinetic features. The event descriptor is
encoded into $e_t(a)$ and scored against the restored state:
\begin{equation}
s_t(a)=g(z_t,e_t(a)).
\label{eq:event_score}
\end{equation}
This memory-conditioned scoring is essential: two events with similar local descriptors
may have different long-horizon consequences depending on the hidden evolutionary
context stored in $z_t$. To preserve physical grounding, the learned score is fused with
KMC and energetic priors:
\begin{equation}
\ell_t(a)
=
\alpha s_t(a)
+
\beta\log r_t^{\mathrm{base}}(a)
+
\eta q_t(a),
\label{eq:event_logit_fusion}
\end{equation}
where $r_t^{\mathrm{base}}(a)$ is the base KMC rate and $q_t(a)$ is an energy-aware
prior. These scores only prioritize among already legal KMC events; they do not
generate unconstrained atomic moves.

\paragraph{Physics-Guided Policy Learner.}
The policy induced by Eq.~\eqref{eq:event_logit_fusion} is trained with PPO using the
shaped reward in Eq.~\eqref{eq:training_reward}. The clipped actor objective is
\begin{equation}
L_{\mathrm{ppo}}
=
-\mathbb{E}_t
\left[
\min
\Big(
\rho_t(\theta)\hat{A}_t,\,
\mathrm{clip}(\rho_t(\theta),1-\epsilon,1+\epsilon)\hat{A}_t
\Big)
\right],
\label{eq:ppo_loss}
\end{equation}
where
\begin{equation}
\rho_t(\theta)=
\frac{
\pi_\theta(a_t\mid z_t,\mathcal{A}^{\mathrm{legal}}(x_t))
}{
\pi_{\theta_{\mathrm{old}}}(a_t\mid z_t,\mathcal{A}^{\mathrm{legal}}(x_t))
}.
\label{eq:ppo_ratio}
\end{equation}
We further use a soft physics teacher $p_t^{\mathrm{phys}}(a)$ derived from explicit
kinetic and energetic cues:
\begin{equation}
L_{\mathrm{phys}}
=
-\mathbb{E}_t
\sum_{a\in\mathcal{A}^{\mathrm{legal}}(x_t)}
p_t^{\mathrm{phys}}(a)
\log
\pi_\theta(a\mid z_t,\mathcal{A}^{\mathrm{legal}}(x_t)).
\label{eq:physics_teacher_loss}
\end{equation}
The full actor objective is
\begin{equation}
L_{\mathrm{actor}}
=
L_{\mathrm{ppo}}
+
\lambda_{\mathrm{phys}}L_{\mathrm{phys}}
+
\lambda_{\mathrm{time}}L_{\mathrm{time}}
-
\lambda_{\mathrm{ent}}
H[\pi_\theta(\cdot\mid z_t)].
\label{eq:actor_loss}
\end{equation}
Here $L_{\mathrm{time}}$ is the restored-state time-consistency objective in
Eq.~\eqref{eq:time_loss}. The critic is also conditioned on $z_t$, so value estimation is
based on the memory-restored evolutionary context rather than on the instantaneous
snapshot alone.

\subsection{Restored-State Time Consistency}

AtomWorld-Mem does not replace the KMC physical clock. During execution, time is
advanced by the standard residence-time rule from the base KMC rates:
\begin{equation}
\Gamma_t(x_t)=
\sum_{a\in\mathcal{A}^{\mathrm{legal}}(x_t)}
r_t^{\mathrm{base}}(a),
\qquad
\Delta t_t^{\mathrm{KMC}}
=
-\frac{\log u_t}{\Gamma_t(x_t)}.
\label{eq:kmc_clock}
\end{equation}
Thus, the time unit, temperature dependence, attempt frequency, and barrier-to-rate
conversion remain those of the underlying simulator. The learned policy changes which
legal event is selected, not how the residence-time increment is computed at a visited
state. We do not claim that AtomWorld-Mem is an exact kinetics-preserving replacement
for passive KMC: the biased trajectory distribution is not guaranteed to satisfy detailed
balance or the stationary path measure of unbiased KMC. Our goal is rare-event discovery
and low-energy structural evolution under legal-event constraints, while monitoring
whether the induced trajectories remain physically meaningful.

Time prediction is induced indirectly through restored-state event hazards rather than
by an independent clock head. We interpret the fused score in
Eq.~\eqref{eq:event_logit_fusion} as an event-wise log-hazard:
\begin{equation}
\lambda_\theta(a\mid z_t)=\exp(\ell_t(a)),
\qquad
\Lambda_\theta(z_t)
=
\sum_{a\in\mathcal{A}^{\mathrm{legal}}(x_t)}
\lambda_\theta(a\mid z_t).
\label{eq:learned_hazard}
\end{equation}
When clear, we suppress the dependence of $\Lambda_\theta$ on the current legal event
set. The same hazard field defines both the legal-event policy and the expected
waiting-time readout:
\begin{equation}
\pi_\theta(a\mid z_t)
=
\frac{\lambda_\theta(a\mid z_t)}
{\Lambda_\theta(z_t)},
\qquad
\widehat{\Delta t}_\theta(z_t)=\Lambda_\theta(z_t)^{-1}.
\label{eq:hazard_policy_time}
\end{equation}
Thus, event preference and time-scale prediction are tied to the same memory-restored
state.

We constrain this temporal readout using the base-rate prior, event-distribution
alignment, and the executed KMC waiting time. Let
\begin{equation}
p_t^{\mathrm{base}}(a)=
\frac{r_t^{\mathrm{base}}(a)}{\Gamma_t(x_t)}.
\label{eq:base_rate_distribution}
\end{equation}
The auxiliary event-time consistency loss is
\begin{equation}
L_{\mathrm{time}}
=
D_{\mathrm{KL}}\!\left(
p_t^{\mathrm{base}}(\cdot)
\,\|\,
\pi_\theta(\cdot\mid z_t)
\right)
+
\log\frac{\Gamma_t(x_t)}{\Lambda_\theta(z_t)}
+
\big(\Lambda_\theta(z_t)-\Gamma_t(x_t)\big)\Delta t_t^{\mathrm{KMC}}.
\label{eq:time_loss}
\end{equation}
The KL term softly regularizes event probabilities toward the physical rate structure,
while the second term compares waiting-time statistics induced by the learned total
hazard against the KMC residence-time target. This loss is not a detailed-balance
constraint and does not make the biased policy path identical to passive KMC; it
calibrates and diagnoses whether the restored state captures the local hazard landscape
of the states it visits.

Throughout the paper, high fidelity does not mean exact reproduction of the unbiased
KMC path measure. It refers to fidelity under three operational constraints:
legal-event execution, KMC-grounded residence-time updates at visited states, and
agreement with reference atomistic dynamics across energetic, structural, and
vacancy-transport observables. Temporal fidelity is therefore a stringent probe of
state-restoration quality: a model may improve energy descent through local heuristics,
but matching the waiting-time scale requires recovering the broader legal-event hazard
landscape.
\section{Experiments}

\subsection{Experimental Setup}




\begin{wrapfigure}{r}{0.43\columnwidth}
    \vspace{-0.8em}
    \centering
    \includegraphics[width=\linewidth]{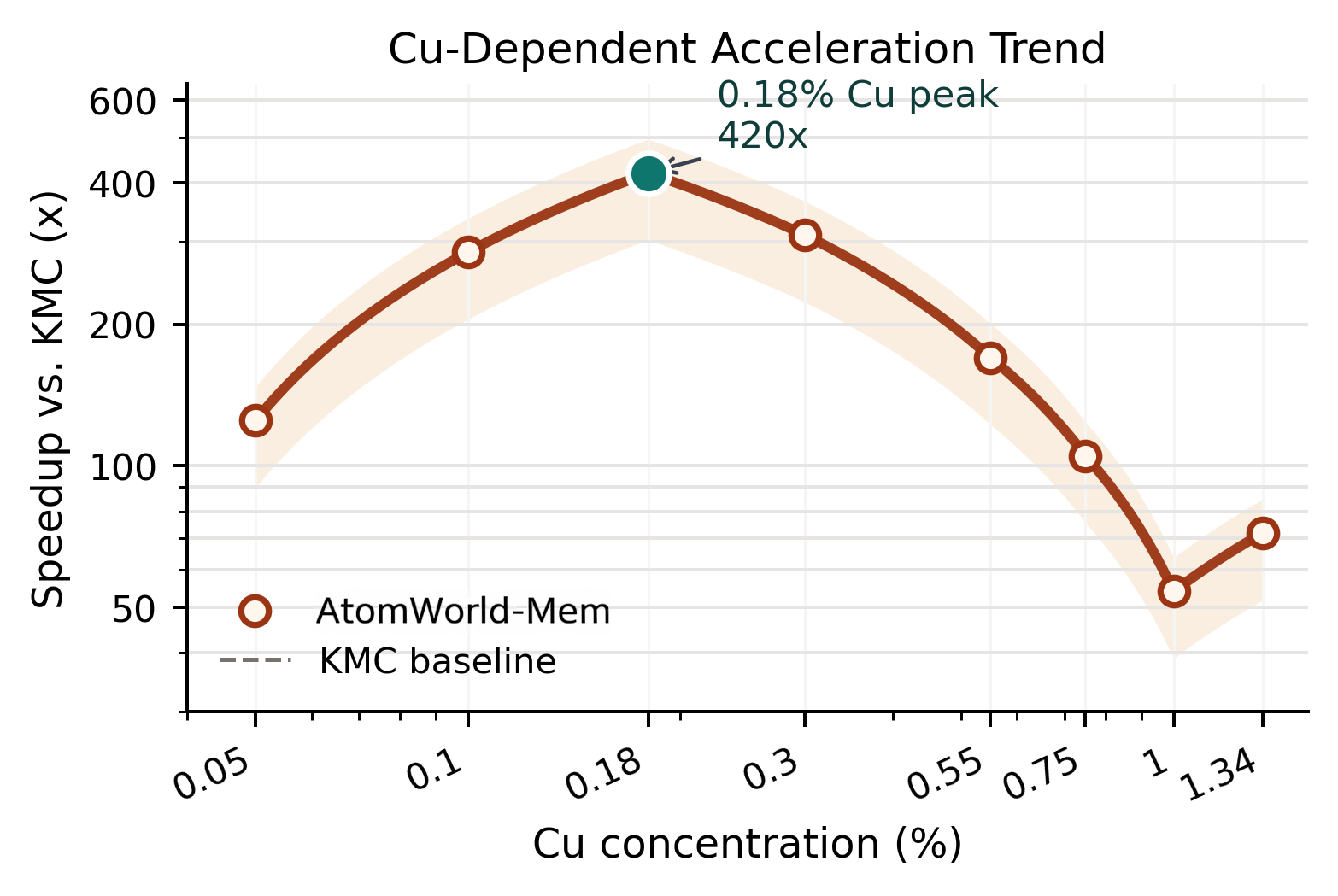}
    \vspace{-0.7em}
    \caption{\textbf{Cu-dependent acceleration trend.}
    Gains remain positive across the tested Cu range, peak at 420$\times$ near
    0.18\% Cu, and average 186$\times$.}
    \label{fig:cu_speedup}
    \vspace{-1.0em}
\end{wrapfigure}

We train and evaluate AtomWorld-Mem on diffusion-driven Fe--Cu alloy evolution,
a canonical vacancy-mediated setting for precipitation and defect-driven structural change.
Each \emph{AtomWorld} has microscopic states that determine legal events, base KMC rates,
and transitions, while the model uses structured snapshots with internal memory.
The actions are legal vacancy-mediated exchanges, and comparisons are step-controlled
unless otherwise stated: methods start from matched initial-state families and use equal
legal-event budgets, isolating fixed-budget event-prioritization quality. Learning
experiments use NVIDIA A100 hardware.

\subsection{Evaluating Memory-Restored AtomWorld Modeling}


Our experiments test whether memory restores latent AtomWorld state by evaluating
fixed-budget progress, trajectory-level fidelity, memory ablations, and zero-shot transfer.

\subsubsection{Fixed-Budget Atomistic Progress}
\begin{table*}[t]
\centering
\caption{
\textbf{Main results under matched legal-event budgets.}
AF is the fixed-budget acceleration factor over Zacros (traditional KMC)~\cite{stamatakis2011graph}, not a claim of
path-measure-preserving kinetic speedup. Agg. Degree measures Cu aggregation, Arrival
Hit measures vacancy arrival to aggregation-relevant Cu clusters, and DynErr measures
deviation from reference KMC dynamics.
}
\label{tab:main_results}

\vspace{0.35em}

\small
\setlength{\tabcolsep}{3.6pt}
\renewcommand{\arraystretch}{1.08}

\resizebox{\linewidth}{!}{
\begin{tabular}{lccccccccc}

\toprule

\textbf{Method}
& \textbf{Memory}
& \multicolumn{4}{c}{\textbf{Dilute setting: 0.18\% Cu}}
& \multicolumn{4}{c}{\textbf{Relatively concentrated setting: 1.34\% Cu}} \\

\cmidrule(lr){3-6}
\cmidrule(lr){7-10}

&
& \makecell{\textbf{AF} $\uparrow$}
& \makecell{\textbf{Agg.}\\\textbf{Degree} $\uparrow$}
& \makecell{\textbf{Arrival}\\\textbf{Hit} $\uparrow$}
& \makecell{\textbf{DynErr} $\downarrow$}
& \makecell{\textbf{AF} $\uparrow$}
& \makecell{\textbf{Agg.}\\\textbf{Degree} $\uparrow$}
& \makecell{\textbf{Arrival}\\\textbf{Hit} $\uparrow$}
& \makecell{\textbf{DynErr} $\downarrow$} \\

\midrule

Zacros (Traditional KMC)
& \xmark
& 1.00$\times$ & 0.016 & 5.0\%  & 0.000
& 1.00$\times$ & 0.072 & 43.8\% & 0.000 \\

Energy-Greedy KMC
& \xmark
& 11.03$\times$ & 0.035 & 45.0\% & 0.466
& 2.16$\times$  & 0.074 & 52.5\% & 0.371 \\

Snapshot-only AtomWorld
& \xmark
& 66.19$\times$ & 0.116 & 67.8\% & 0.318
& 11.13$\times$ & 0.166 & 67.5\% & 0.299 \\

\rowcolor{AtomWorldRow}
\textbf{AtomWorld-Mem}
& \cmark
& \textbf{420.05$\times$}
& \textbf{0.432}
& \textbf{85.5\%}
& \textbf{0.221}
& \textbf{51.25$\times$}
& \textbf{0.367}
& \textbf{80.9\%}
& \textbf{0.262} \\

\bottomrule

\end{tabular}
}

\end{table*}

\begin{figure*}[t]
    \centering
    \includegraphics[width=\textwidth]{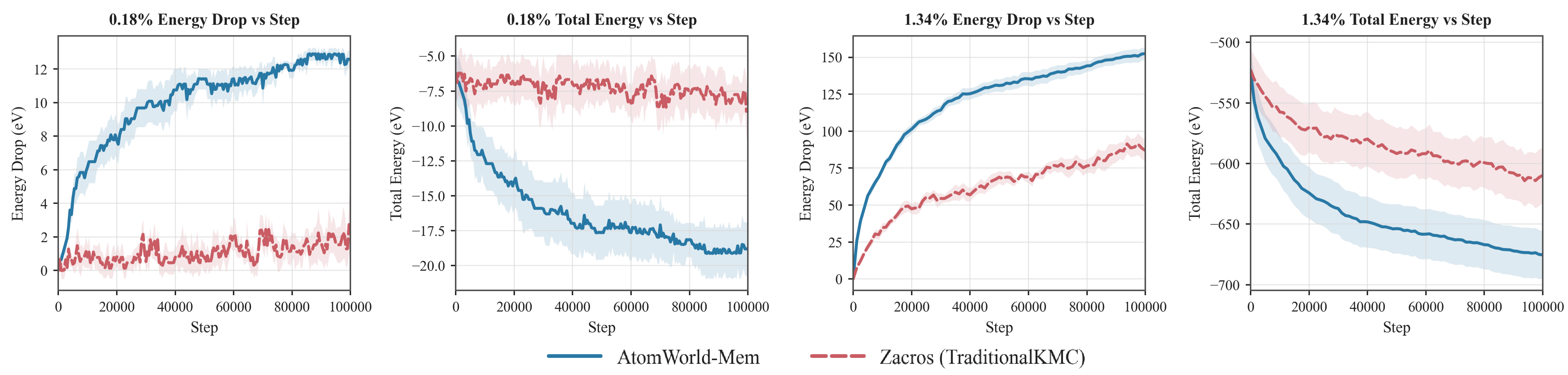}
    \caption{
    \textbf{Fixed-budget energy evolution over 100k legal KMC steps.}
    From left to right: energy drop and total energy at 0.18\% Cu, followed by energy
    drop and total energy at 1.34\% Cu. AtomWorld-Mem achieves larger energy drops and
    lower total energies than Zacros under the same legal-event budget. Shaded regions
    indicate trajectory variability across runs.
    }
    \label{fig:energy_compare}
\end{figure*}

We first test whether restoring latent evolutionary context improves use of a fixed
event budget. All methods start from matched initial-state families and execute the
same number of legal vacancy-mediated events, so differences reflect event
prioritization rather than rollout length.



Table~\ref{tab:main_results} separates passive sampling, local energy descent,
snapshot-based spatial reasoning, and memory-restored state inference. Snapshot-only
AtomWorld already improves over Zacros, showing that spatial world representations
contain useful event context. However, adding memory yields the decisive gain: AF rises
from 66.19$\times$ to 420.05$\times$ at 0.18\% Cu and from 11.13$\times$ to
51.25$\times$ at 1.34\% Cu. AtomWorld-Mem also improves Arrival Hit and DynErr,
indicating that memory improves not only scalar progress but also long-horizon event
prioritization quality.


Figure~\ref{fig:energy_compare} shows that AtomWorld-Mem separates early from Zacros
and maintains its advantage throughout the 100k-step rollout, ruling out an
endpoint-only gain. The restored state repeatedly reallocates legal events toward
structurally productive transitions. Figure~\ref{fig:cu_speedup} further shows that
the advantage persists across Cu concentrations, averaging 186$\times$. These results
support the first prediction of our hypothesis: hidden-state restoration improves
fixed-budget atomistic progress beyond snapshot-level event scoring or local energy
heuristics.


\subsubsection{Restored-State Fidelity and Correctness}

Fast progress alone does not imply a high-fidelity atomistic world model: a policy may
accelerate energy descent by exploiting local shortcuts. We therefore use
trajectory-level fidelity as a stricter test of state restoration. Since AtomWorld-Mem
intentionally biases legal-event selection, we do not claim exact reproduction of the
unbiased KMC path measure or detailed balance. Instead, high fidelity here means that
the induced rollout remains aligned with reference atomistic dynamics across energetic,
structural, and vacancy-transport observables.

\begin{table}[t]
\centering
\caption{
\textbf{Memory ablation at 0.18\% Cu.}
Short/Long denote short-term event memory and long-term structural memory. TimeErr and
DynErr measure temporal and rollout deviation from Zacros.
}
\label{tab:memory_ablation_018}

\vspace{0.2em}

\footnotesize
\setlength{\tabcolsep}{2.4pt}
\renewcommand{\arraystretch}{0.92}

\resizebox{\linewidth}{!}{
\begin{tabular}{lccccccc}

\toprule

\textbf{Variant}
& \textbf{Short}
& \textbf{Long}
& \textbf{AF}$\uparrow$
& \textbf{Agg.}$\uparrow$
& \textbf{TimeErr}$\downarrow$
& \textbf{DynErr}$\downarrow$
& \textbf{Arr.Hit}$\uparrow$ \\

\midrule

Zacros (Traditional KMC)
& \xmark & \xmark
& 1.00$\times$
& 0.016
& 0.000
& 0.000
& 5.00\% \\

Snapshot-only AtomWorld
& \xmark & \xmark
& 62.16$\times$
& 0.089
& 0.777
& 0.591
& 65.30\% \\

No short-term memory
& \xmark & \cmark
& 127.43$\times$
& 0.183
& 0.475
& 0.293
& 71.64\% \\

No long-term memory
& \cmark & \xmark
& 148.25$\times$
& 0.235
& 0.599
& 0.348
& 73.91\% \\

\rowcolor{AtomWorldRow}
\textbf{Full AtomWorld-Mem}
& \cmark & \cmark
& \textbf{418.44$\times$}
& \textbf{0.426}
& \textbf{0.308}
& \textbf{0.192}
& \textbf{83.22\%} \\

\bottomrule

\end{tabular}
}

\vspace{-0.8em}

\end{table}

\begin{figure*}[t]
    \centering
    \includegraphics[width=\textwidth]{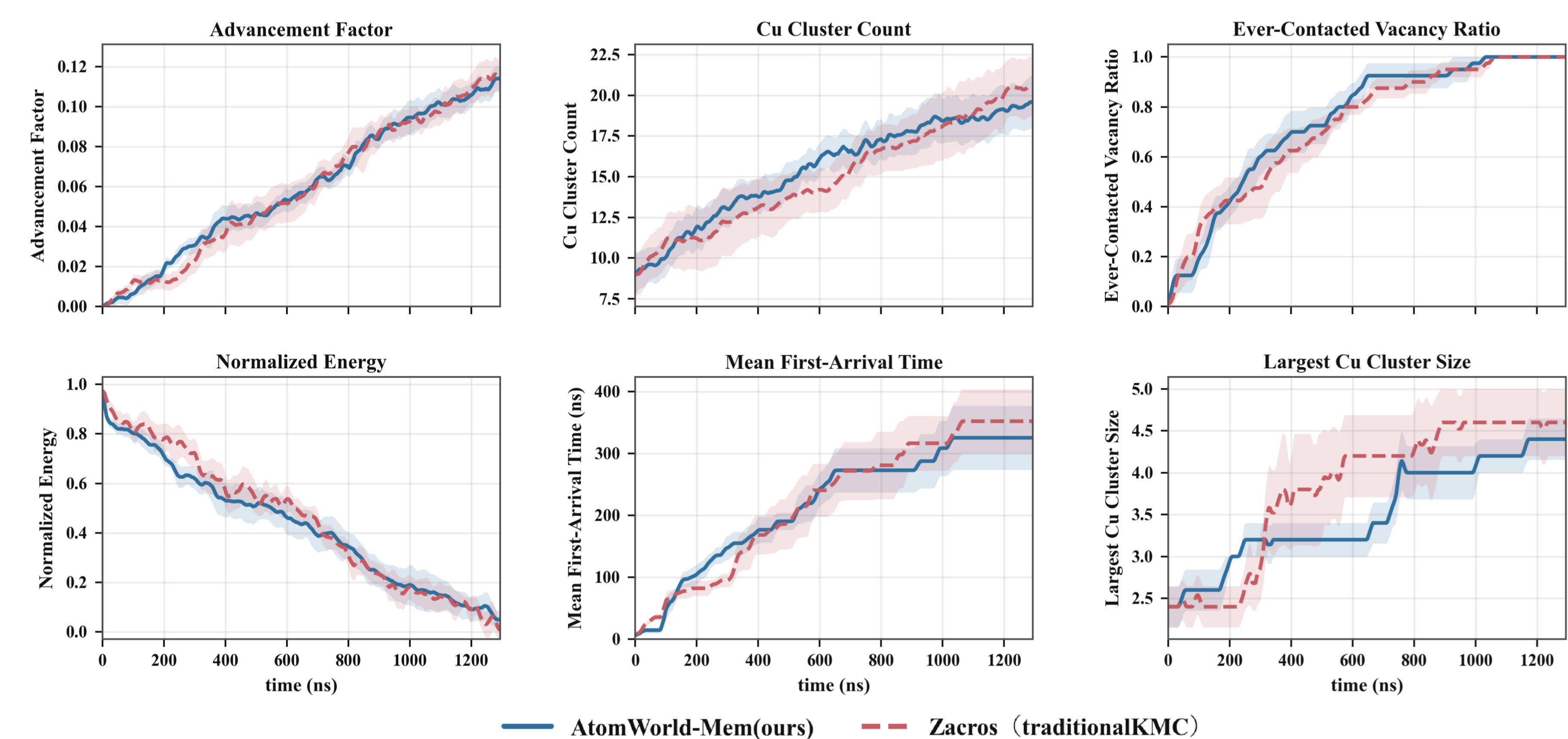}
    \caption{
    \textbf{High-fidelity atomistic evolution in physical-time space.}
    We compare AtomWorld-Mem with Zacros traditional KMC across six trajectory-level
    observables: advancement factor, Cu cluster count, ever-contacted vacancy ratio,
    normalized energy, mean first-arrival time, and largest Cu cluster size.
    AtomWorld-Mem follows the reference trends while maintaining faster fixed-budget
    progress, indicating that memory-restored event prioritization does not collapse
    into an unphysical local shortcut.
    }
    \label{fig:correctness_validation}
\end{figure*}

Figure~\ref{fig:correctness_validation} shows that AtomWorld-Mem remains broadly
aligned with Zacros in physical-time space. Normalized energy follows the same
relaxation trend; cluster count and largest-cluster size track the aggregation regime;
and vacancy-related metrics preserve the ordering and timing of vacancy--cluster
interactions. These observables test fidelity beyond scalar energy descent. The result
supports the restored-state interpretation: AtomWorld-Mem improves long-horizon
event prioritization because memory recovers evolutionary context missing from
instantaneous snapshots, rather than because the policy exploits a system-specific
local energy shortcut.

\subsubsection{Memory Ablation: Does Memory Restore the Missing State?}

We next test the mechanism directly: if memory restores hidden evolutionary state,
removing either memory scale should degrade both progress and fidelity.
Table~\ref{tab:memory_ablation_018} shows that snapshot-only spatial reasoning is useful
but incomplete, with improved AF but large TimeErr and DynErr. Adding either memory
scale improves progress and fidelity, while the full model performs best across learned
variants, increasing AF to 418.44$\times$ and reducing DynErr to 0.192. This indicates
that AtomWorld-Mem benefits from temporal state restoration rather than only a stronger
instantaneous event scorer.

\subsubsection{Zero-Shot Transfer Across AtomWorlds}

\begin{figure*}[t]
    \centering
    \includegraphics[width=\textwidth]{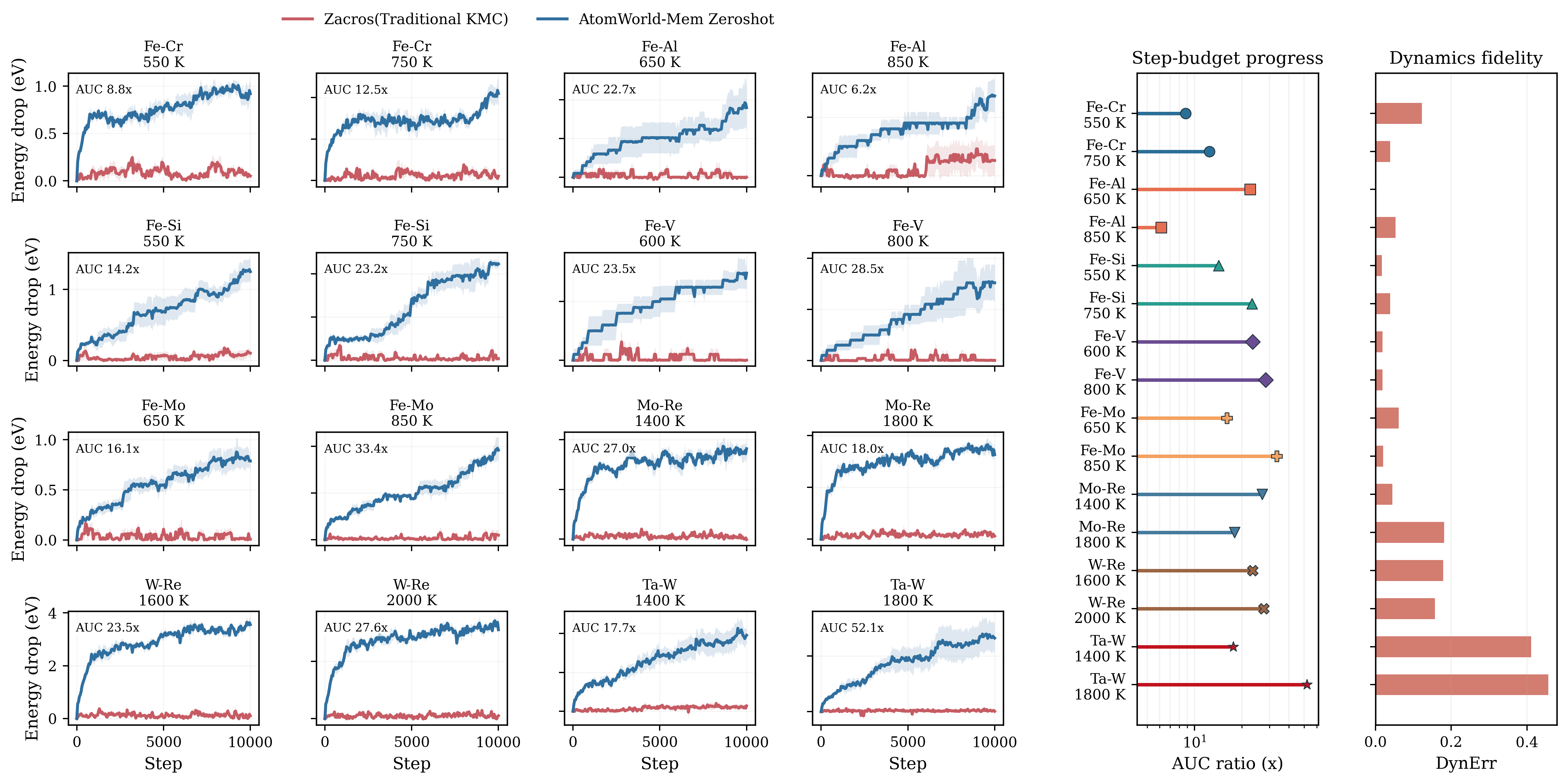}
    \caption{
    \textbf{Zero-shot transfer across 16 unseen BCC alloy--temperature AtomWorlds.}
    Left: energy-drop trajectories under a shared 10k-step budget. Right: per-domain
    AUC acceleration ratio and dynamics-fidelity error.
    }
    \label{fig:cross_alloy_zero_shot}
\end{figure*}

We finally evaluate zero-shot transfer to unseen alloy--temperature AtomWorlds without
finetuning. We evaluate 16 unseen BCC alloy--temperature domains spanning Fe--Cr, Fe--Al, Fe--Si,
Fe--V, Fe--Mo, Mo--Re, W--Re, and Ta--W. As shown in
Figure~\ref{fig:cross_alloy_zero_shot}, AtomWorld-Mem improves fixed-budget energy-drop
trajectories in every domain, with a median AUC gain of \textbf{22.9$\times$} and a
maximum gain of \textbf{52.1$\times$}. Dynamics-fidelity error remains small or moderate
for most domains, with larger deviations mainly in refractory systems.
These consistent gains suggest transferable memory-based state inference rather than an
Fe--Cu-specific local energy heuristic.

\section{Related Work}

\paragraph{Kinetic Monte Carlo and policy-guided atomistic kinetics.}
Kinetic Monte Carlo (KMC) simulates long-timescale atomistic evolution through
physically admissible events and residence-time updates~\cite{gillespie1977exact, gillespie2007stochastic, voter2007introduction, fichthorn1991theoretical, pineda2022kinetic, andersen2019practical}. 
Adaptive, accelerated,
self-evolving, ML-assisted, and policy-guided variants improve event discovery,
local updates, rate estimation, or transition selection under physical constraints~\cite{henkelman2001long, xu2008adaptive, pedersen2010distributed, trushin2005self, mousseau2008kinetic, messina2017introducing, kimari2020application, tang2024reinforcement}.
However, these methods typically condition decisions on the current local configuration
or emphasize aggregate sampling efficiency, leaving \emph{state incompleteness} largely
unaddressed: similar snapshots may hide different dynamical contexts and long-horizon
event preferences~\cite{trushin2005self, xu2008adaptive, chatterjee2010accurate}. 
AtomWorld-Mem retains KMC-grounded execution and timing, but uses
memory to restore a future-predictive state for legal-event prioritization~\cite{gillespie1977exact, fichthorn1991theoretical}.

\paragraph{World models and latent dynamics under partial observability.}
World models and latent-dynamics methods learn compact internal states from partial
observations for prediction, planning, and control~\cite{hafner2019learning, karl2016deep, buesing2018learning, ha2018recurrent, hafner2023mastering}. They often learn or imagine
environment transitions in latent space, whereas atomistic KMC already provides a
physically constrained transition system~\cite{hafner2019dream, hafner2020mastering, fichthorn1991theoretical, andersen2019practical}. AtomWorld-Mem adopts the latent-state view
for a different purpose: restoring hidden dynamical context to rank admissible KMC
events, while leaving event legality, execution, and residence-time updates to the
underlying simulator~\cite{littman2001predictive, hefny2018recurrent, meng2021memory, voter2007introduction}.

\section{Conclusion and Limitations}


We presented AtomWorld-Mem, a memory-restored atomistic world model for
long-horizon evolution under physical constraints. The key idea is to treat
atomistic evolution as hidden-state restoration rather than snapshot-level event
sampling. Across progress, fidelity, ablation, and transfer tests, memory-restored
states outperform snapshot-only reasoning while remaining physically grounded.

This work has limitations. Our experiments focus on vacancy-mediated BCC alloy evolution; extending the framework to richer materials, more complex event catalogs, stronger clock-consistency objectives, and more expressive memory mechanisms remains an important direction. Overall, our results position memory-restored world-state modeling as a promising route toward efficient, physically grounded, and transferable atomistic evolution.




\newpage
{\small
\bibliographystyle{unsrtnat}
\bibliography{references}
}

\newpage
\appendix

\section{Experimental Details}
\subsection{Simulation Environments}

The main experiments use vacancy-mediated Fe--Cu alloy evolution, a canonical setting
for precipitation and defect-driven structural change. Each environment is treated as an
AtomWorld: the full microscopic configuration determines legal events, base KMC rates,
and physical transitions, while the learned model receives only structured partial
observations and memory. The action space consists of vacancy-mediated atomic exchanges
with local-environment-dependent Arrhenius rates.

Unless otherwise stated, comparisons are step-controlled. All methods start from matched
initial-state families and execute the same number of legal KMC events. This isolates
event-prioritization quality: differences in progress reflect how effectively a method
uses the same microscopic event budget, rather than differences in rollout length.

\subsection{Baselines and Ablations}

\paragraph{Zacros traditional KMC.}
Zacros traditional KMC is the passive reference simulator. It samples legal events
proportionally to their base transition rates and advances physical time using the
standard residence-time rule. It provides the reference for both fixed-budget progress
and trajectory-level fidelity.

\paragraph{Energy-Greedy KMC.}
Energy-Greedy KMC selects legal events using local energetic improvement. This baseline
tests whether the gains of AtomWorld-Mem can be explained by a simple local
energy-descent heuristic.

\paragraph{Snapshot-only AtomWorld.}
Snapshot-only AtomWorld keeps the instantaneous spatial representation but removes
temporal memory. This baseline tests whether multi-scale spatial observations alone are
sufficient, or whether hidden evolutionary context must be restored from history.

\paragraph{No short-term memory.}
This variant removes recent-event memory while retaining long-term structural memory.
It tests the role of recent action history, local transition competition, vacancy
backtracking, and short-horizon event reasoning.

\paragraph{No long-term memory.}
This variant removes recurrent structural memory while retaining short-term event
history. It tests the role of slow structural-context accumulation, hidden
defect-evolution trends, and long-horizon aggregation bias.

\section{Additional Results}

\subsection{Additional Memory Ablations}

The main text reports memory ablations in the dilute regime (0.18\% Cu). Here we
further provide the corresponding ablation in a relatively concentrated setting
(1.34\% Cu) to test whether the role of memory persists when the alloy exhibits richer
local competition and stronger structural interactions.

\begin{table}[!htbp]
\centering
\caption{
\textbf{Additional memory ablation at 1.34\% Cu.}
Short/Long denote short-term event memory and long-term structural memory. AF is the
step-to-target acceleration factor over Zacros traditional KMC. TimeErr is log expected
waiting-time MAE; DynErr is real-KMC-time aligned NRMSE to Zacros.
}
\label{tab:memory_ablation_134}
\vspace{0.25em}
\footnotesize
\setlength{\tabcolsep}{2.8pt}
\renewcommand{\arraystretch}{0.94}
\resizebox{\linewidth}{!}{
\begin{tabular}{lccccccc}
\toprule
\textbf{Variant}
& \textbf{Short}
& \textbf{Long}
& \textbf{AF}$\uparrow$
& \textbf{Agg.}$\uparrow$
& \textbf{TimeErr}$\downarrow$
& \textbf{DynErr}$\downarrow$
& \textbf{Arr.Hit}$\uparrow$ \\
\midrule
Snapshot-only AtomWorld
& \xmark & \xmark
& 11.35$\times$ & 0.169 & 0.342 & 0.332 & 68.80\% \\
No short-term memory
& \xmark & \cmark
& 21.33$\times$ & 0.202 & 0.219 & 0.206 & 70.85\% \\
No long-term memory
& \cmark & \xmark
& 28.36$\times$ & 0.257 & 0.255 & 0.282 & 75.71\% \\
\textbf{Full AtomWorld-Mem}
& \cmark & \cmark
& \textbf{51.38$\times$} & \textbf{0.358} & \textbf{0.141} & \textbf{0.165} & \textbf{83.82\%} \\
\bottomrule
\end{tabular}
}
\vspace{-0.6em}
\end{table}

Table~\ref{tab:memory_ablation_134} shows the same qualitative pattern as the dilute
regime, confirming that the benefit of memory is not concentration-specific.
Snapshot-only spatial reasoning already improves over passive sampling, but remains
substantially weaker than the full model. Adding either memory scale improves both
progress and fidelity, while the full AtomWorld-Mem performs best across all metrics.

The ablation also clarifies the complementary roles of the two memory scales. Removing
short-term memory primarily weakens recent-event reasoning and local transition
competition, which is reflected in lower vacancy-arrival quality and reduced progress.
Removing long-term memory more strongly degrades structural-context accumulation and
rollout-level consistency, leading to weaker aggregation and larger dynamics error.
The full model combines both: short-term memory supports local temporal reasoning,
whereas long-term memory preserves slow structural context across the trajectory.
Together, these results further support the state-restoration hypothesis: the advantage
of AtomWorld-Mem comes from recovering hidden evolutionary state across time, rather
than from a stronger snapshot-level event scorer.

\subsection{Long-Horizon Visualization Details}

We further visualize long-horizon structural evolution to verify that accelerated
progress corresponds to physically meaningful mesoscale organization rather than only
improved scalar energy descent.

\begin{figure}[!htbp]
    \centering
    \includegraphics[width=\linewidth]{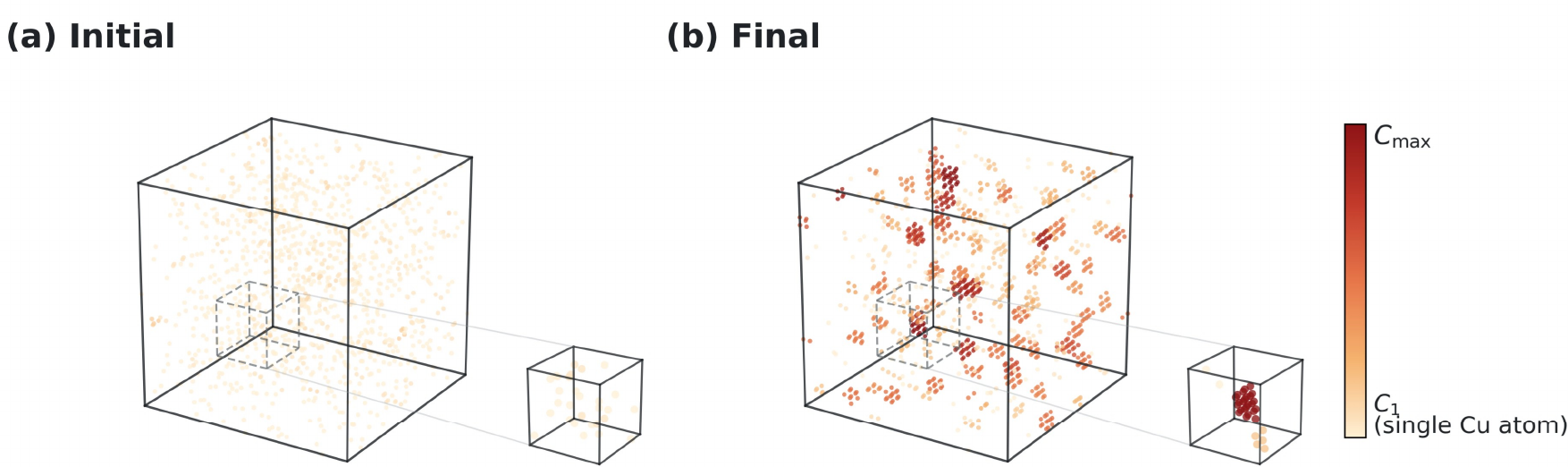}
    \caption{
    \textbf{Long-horizon Cu clustering visualization.}
    Cu clustering evolution in Fe--0.67 at.\% Cu over a 50-year physical-time window.
    The visualization illustrates whether memory-restored event prioritization produces
    coherent long-horizon structural organization rather than only faster energy
    reduction.
    }
    \label{fig:long_horizon_visualization}
    \vspace{-0.6em}
\end{figure}

Figure~\ref{fig:long_horizon_visualization} provides a qualitative complement to the
quantitative metrics in the main text. Trajectories are rendered at matched
physical-time anchors, with fixed visualization settings across methods. Cu atoms are
grouped into clusters using the same neighborhood-based rule used for aggregation
metrics, so that visual differences reflect trajectory behavior rather than rendering
choices.

The visualization supports the same conclusion as the trajectory-level metrics:
AtomWorld-Mem does not merely exploit a local energy shortcut. Instead, the
memory-restored state guides legal vacancy-mediated events toward coherent clustering
and sustained long-horizon structural evolution. In this view, short-term memory helps
capture recent vacancy--cluster interactions, while long-term memory preserves the slow
structural context needed for persistent aggregation over physical time.

\end{document}